\documentclass[journal]{IEEEtran}
\usepackage{cite}
\usepackage{amsmath,amssymb,amsfonts}
\usepackage{algorithmic}
\usepackage{graphicx}
\usepackage{textcomp}

\usepackage{latexsym}
\usepackage{booktabs}
\usepackage{pgfplots}
\usepackage{multirow}
\usepackage{float}
\usepackage{mathrsfs}
\usepackage[ruled]{algorithm2e}
\usepackage{etoolbox}

\usepackage{soul}
\usepackage{xcolor}
\usepackage{tabularray}
\usepackage{lipsum}
\usepackage{xr}
\usepackage{comment}
\usepackage{subfigure}
\usepackage{hhline}
\usepackage{makecell}
\usepackage{wrapfig}
\usepackage{bm}
\usepackage[hidelinks]{hyperref}

\providecommand{\cite}[1]{\cite{#1}}

\def\BibTeX{{\rm B\kern-.05em{\sc i\kern-.025em b}\kern-.08em
    T\kern-.1667em\lower.7ex\hbox{E}\kern-.125emX}}
\begin{document}

\title{Structured Noise Diffusion for Echocardiography Haze Removal}
\author{Zike Luo$^\dagger$, Hao Li$^\dagger$, Mengxiao Wang, Qingya Li, Ching-Hui Sia, and Lei Li*
\thanks{Corresponding author: L. Li (e-mail: lei.li@nus.edu.sg). 
This work was supported by Singapore National Medical Research Council Open Fund - Young Individual Research Grant (MOH-001977-01), and Singapore Ministry of Education Tier 1 grant (25-1097-P0001). 
}
\thanks{Z. Luo, H. Li, M. Wang, Q. Li, and L. Li are with Department of Biomedical Engineering, National University of Singapore, Singapore. M. Wang is also with School of Mechanical Engineering, Shanghai Jiao Tong University, Shanghai, China.}
\thanks{C.H. Sia is with the Department of
Cardiology, National University Heart Centre Singapore, Singapore and the Department of Medicine,
National University of Singapore, Singapore.}
\thanks{$^\dagger$Z. Luo and H. Li contributed equally to this work.}
}

\maketitle

\begin{abstract}

Echocardiographic images are frequently degraded by spatially structured haze artifacts that obscure anatomical boundaries and hinder clinical analysis. Conventional diffusion models typically employ pixel-independent Gaussian noise, limiting their ability to represent localized ultrasound haze. In this study, we propose a structure-informed diffusion framework that explicitly models ultrasound haze as an accumulation of localized Gaussian components constrained by the ultrasound fan geometry. The forward process generates spatially correlated corruption by sampling component locations and scales, providing a formulation that better reflects the spatial characteristics of real echo artifacts. During reverse restoration, a dual-head network predicts an anchor probability map and Gaussian scale parameters to progressively remove structured artifacts while reducing the center bias associated with direct coordinate regression. We evaluated the framework on the public DehazingEcho2025 benchmark and two private clinical datasets, without retraining or fine-tuning on either private cohort. Our method achieved the highest mean generalized contrast-to-noise ratio across all three datasets, reaching 0.715 ± 0.090 on the public benchmark while preserving tissue edges. 
\end{abstract}


\section{Introduction}

\begin{figure}[t] 
  \centering
   \includegraphics[width=1\linewidth]{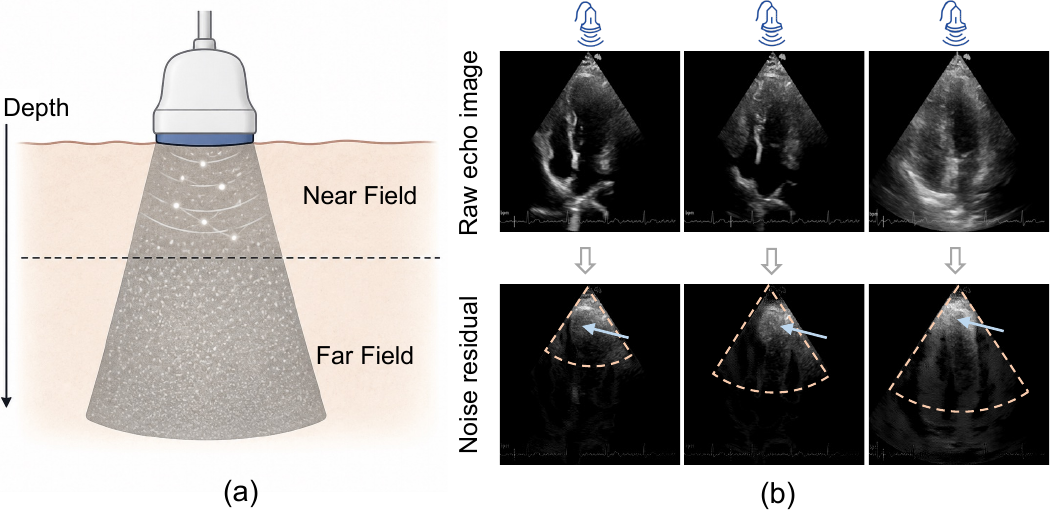}
   \caption{(a) Illustration of the near-field and far-field regions in ultrasound imaging. (b) Raw echo images and corresponding noise residual obtained by subtracting the denoised image from the raw image.}
   \label{fig:intro:noise}
\end{figure}

Echocardiography (echo) is a non-invasive, real-time, and low-cost imaging modality that plays an indispensable role in the clinical diagnosis of cardiovascular diseases~\cite{journal/MedIA/li2023}. However, due to the nature of ultrasound imaging, acquired images commonly suffer from quality degradation in the form of noise, artifacts, and blurring. Among these, haze artifacts are particularly problematic, further impairing the ability to identify anatomical structures and make accurate diagnostic decisions~\cite{journal/UMB/Fatemi2019,journal/ACMP/Sass2019,journal/FCM/Chen2020}.
Unlike atmospheric haze in natural images, which tends to be spatially uniform, haze in ultrasound images exhibits pronounced spatial heterogeneity~\cite{journal/TMI/Stevens2024}. As shown in Fig.~\ref{fig:intro:noise}, noise residuals are heavily concentrated in the near-field region close to the transducer, while image quality in the far field is generally superior. This near-field dominance is mainly caused by strong acoustic reverberation and multipath reflections generated at shallow tissue interfaces and the probe-skin contact region, where echoes undergo repeated scattering before sufficient depth-dependent attenuation occurs. Therefore, haze artifacts in echo images are inherently localized, depth-dependent, and spatially structured rather than uniformly distributed across the image domain.

\begin{figure*}[t] 
  \centering
   \includegraphics[width=0.72\linewidth]{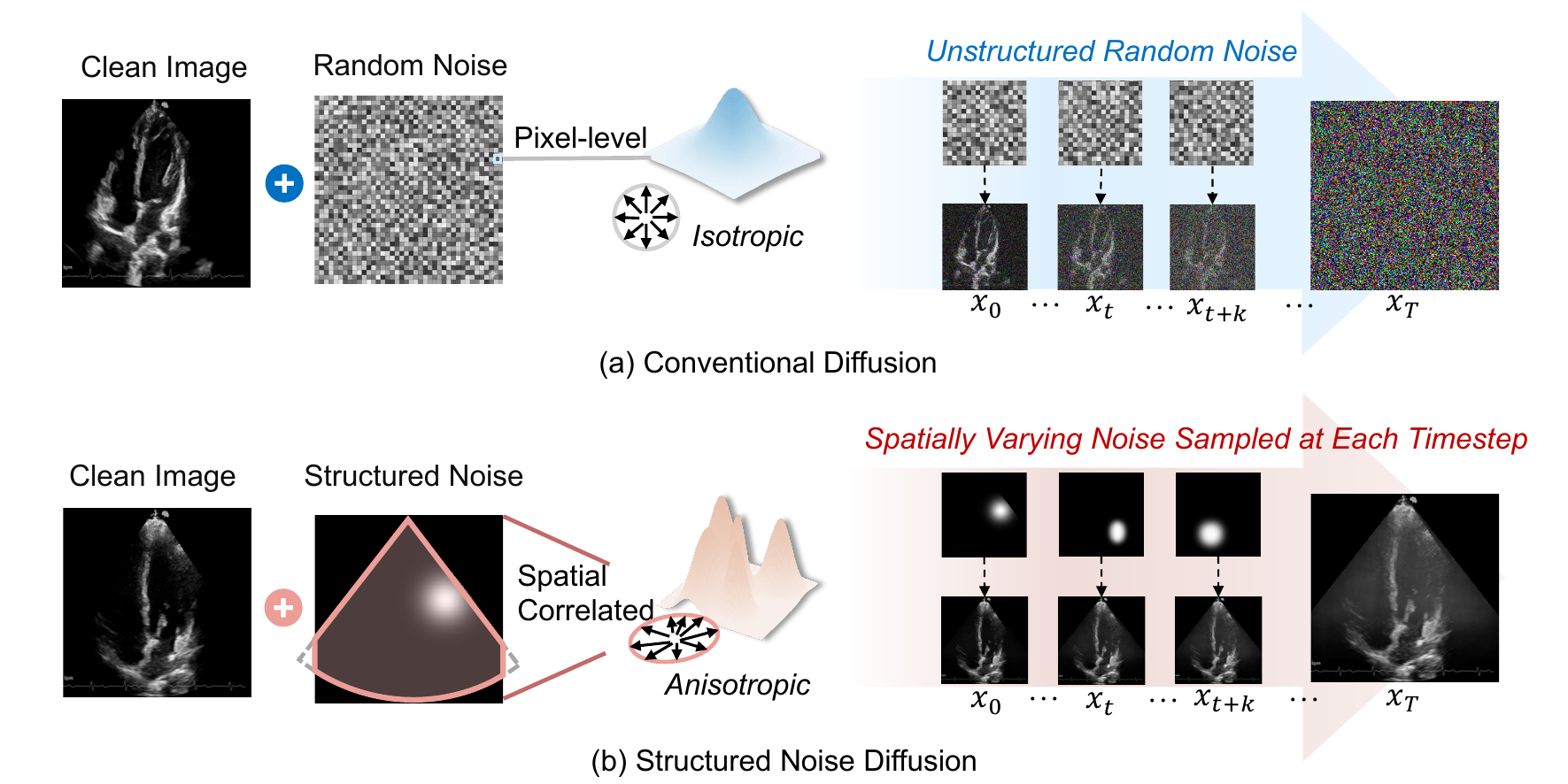}
   \caption{Comparison between conventional diffusion and the proposed structured noise diffusion. For conventional diffusion, unstructured gaussian noise randomly adds at pixel-level image. For structured noise diffusion, spatially correlated, anisotropic noise is sampled at each timestep within a fan-shaped valid echo region.}
   \label{fig:intro:teaser}
\end{figure*}

To address haze artifacts in ultrasound imaging, a range of restoration approaches have been investigated. Conventional methods, such as block-matching and 3D filtering (BM3D)~\cite{conf/SPIE/Dabov2006} and non-local means~\cite{journal/VRIH/Liu2023}, typically formulate dehazing as a signal-statistical denoising problem. These methods exploit local or non-local patch similarity to suppress noise and blur, but their fixed assumptions make it difficult to adapt to the spatially heterogeneous and anatomy-dependent artifacts commonly observed in echo. More recently, deep generative models have shown promise for ultrasound image enhancement. Generative adversarial networks (GANs) based approaches~\cite{journal/IDA/Jim2024,journal/PCS/Zhang2022,journal/IA/Vieira2024} aim to translate degraded images into a cleaner image domain, often improving perceptual quality, but they may hallucinate or over-smooth clinically important structures when the degradation process is not explicitly modeled. Diffusion-based methods~\cite{journal/ArX/Stevens2025,conf/IUS/Asgar2023} provide a more principled iterative restoration framework by learning to recover clean images from progressively corrupted inputs. However, most existing diffusion formulations still rely on globally applied, pixel-independent Gaussian noise, implicitly assuming that degradation is uniformly distributed across the image domain. This assumption is inconsistent with echo haze, which is localized, depth-dependent, and spatially correlated due to acoustic reverberation and multi-path scattering. As a result, uniformly modeled noise may insufficiently represent severe near-field haze while unnecessarily perturbing relatively clean far-field regions, limiting the ability of the model to remove structured artifacts without compromising fine anatomical details.

In this study, we propose a novel structure-informed local diffusion model for echo haze removal, motivated by the observation that ultrasound haze is spatially localized, anisotropic, and constrained by the fan-shaped imaging geometry. As illustrated in Fig.~\ref{fig:intro:teaser}, unlike conventional diffusion models that perturb images with pixel-independent isotropic Gaussian noise, our method represents haze as a sequence of localized 2D Gaussian components sampled within the valid echo region. Each component is defined by a stochastic anchor point and scale parameters, allowing the forward process to mimic spatially varying acoustic clutter while avoiding unrealistic corruption outside the ultrasound fan.
In the reverse process, instead of directly predicting a dense noise map or regressing raw anchor coordinates, the network estimates a probabilistic anchor map together with the corresponding Gaussian noise parameters at each denoising step. This probabilistic formulation reduces the center-bias problem commonly associated with direct coordinate regression and enables the model to adaptively localize diverse haze sources across different images and diffusion timesteps. The predicted parameters are then used to reconstruct the structured haze field, which is iteratively removed to recover the clean echo image. The main contributions of this work are summarized as follows:
\begin{enumerate}
\item We propose a structure-informed diffusion framework for ultrasound haze removal, reformulating echo dehazing as a structured artifact modeling problem rather than conventional pixel-wise noise suppression.
\item We introduce a fan-constrained structured diffusion process based on stochastic localized Gaussian components, providing a more realistic formulation for spatially heterogeneous and depth-dependent ultrasound haze.
\item We design a probabilistic reverse estimation strategy that predicts anchor probability maps and Gaussian noise parameters, enabling spatially adaptive haze removal while avoiding coordinate regression bias.
\item We validate the proposed method on the DehazingEcho2025 dataset, achieving state-of-the-art performance.
\end{enumerate}

\section{Related Work}

\subsection{Ultrasound Image Denoising and Dehazing}

Ultrasound image restoration differs fundamentally from natural image denoising because its degradation is governed by acoustic wave propagation and tissue interaction rather than simple additive sensor noise. Speckle noise, one of the most characteristic ultrasound degradations, is multiplicative and spatially correlated, arising from coherent interference of backscattered echoes~\cite{journal/ESA/Mateo2009}. In addition, ultrasound signals may be represented as raw radio-frequency (RF) data, envelope-detected signals, or log-compressed B-mode images, each exhibiting distinct statistical properties and requiring different modeling assumptions~\cite{journal/IiI/Cloutier2021}. Despite these modality-specific characteristics, many ultrasound restoration methods have followed the general evolution of image restoration techniques, ranging from local statistical filtering and non-local similarity methods to CNNs, GANs, and diffusion models~\cite{journal/IDA/Jim2024,journal/TMI/Stevens2024}.
Several studies have attempted to incorporate ultrasound-specific priors into restoration models. For example, Gaussian-based filtering methods adapted smoothing strength according to locally estimated speckle statistics~\cite{journal/Measurement/Andria2012}, while speckle-reducing anisotropic diffusion reformulated the diffusion coefficient to better respect the multiplicative nature of ultrasound noise~\cite{journal/PR/Yu2010}. More recently, speckle denoising diffusion models simulated multiplicative noise in the forward process to preserve speckle texture while enhancing anatomical structures~\cite{journal/ArX/Guha2023}. Beyond image-domain methods, RF-domain approaches exploited the complex pre-log-compression signal for more principled noise estimation~\cite{journal/JHE/Khavari2018}, and spatio-temporal methods leveraged cardiac cycle consistency to suppress artifacts in echocardiographic sequences~\cite{journal/MIA/Wu2015}. However, existing ultrasound restoration models mainly focus on speckle or generic denoising, while echo haze exhibits a distinct spatial pattern: localized, depth-dependent, and often concentrated within the fan-shaped near-field region. 

\subsection{Structured Noise Diffusion}

Standard diffusion models typically define the forward process using isotropic Gaussian noise and learn a reverse process to recover clean data from progressively corrupted samples. This formulation is powerful and general, but its pixel-independent corruption assumption may be suboptimal for data with strong structural constraints. In natural images, geometry-aware editing and restoration methods have shown that explicitly controlling the noise schedule or sampling path can better preserve layout and boundaries~\cite{journal/ArX/qi2024,journal/ArX/Jiang2026}. Similarly, structured transition designs such as D3PMs extend diffusion modeling beyond continuous Gaussian perturbations by defining structured corruption processes for discrete state spaces~\cite{conf/Nips/austin2021}. These studies suggest that the choice of forward degradation is not merely a technical detail, but an important inductive bias that directly influences the stability and fidelity of reverse generation.
Structured diffusion has also gained attention in medical and scientific imaging, where degradation is often constrained by anatomy, acquisition physics, or domain-specific geometry. For example, anatomy-guided diffusion frameworks incorporate boundary or shape priors to improve lesion reconstruction fidelity~\cite{journal/ArX/Kwak2025}; seismic denoising methods use structure tensors to guide diffusion along coherent signal directions~\cite{journal/TGRS/zhang2019}; and document restoration models employ hierarchical structural priors to recover text and layout information from local to global scales~\cite{journal/SSRN/Lu2025}. Related work has also explored joint modeling of signal and structured noise to separate artifacts more effectively during the reverse process~\cite{journal/TMLR/stevens2025}. Despite these advances, existing structured diffusion methods have not explicitly addressed echo haze, whose degradation is simultaneously fan-constrained, spatially correlated, and acoustically localized. 

\begin{figure*}[t]
  \centering
   \includegraphics[width=0.7\linewidth]{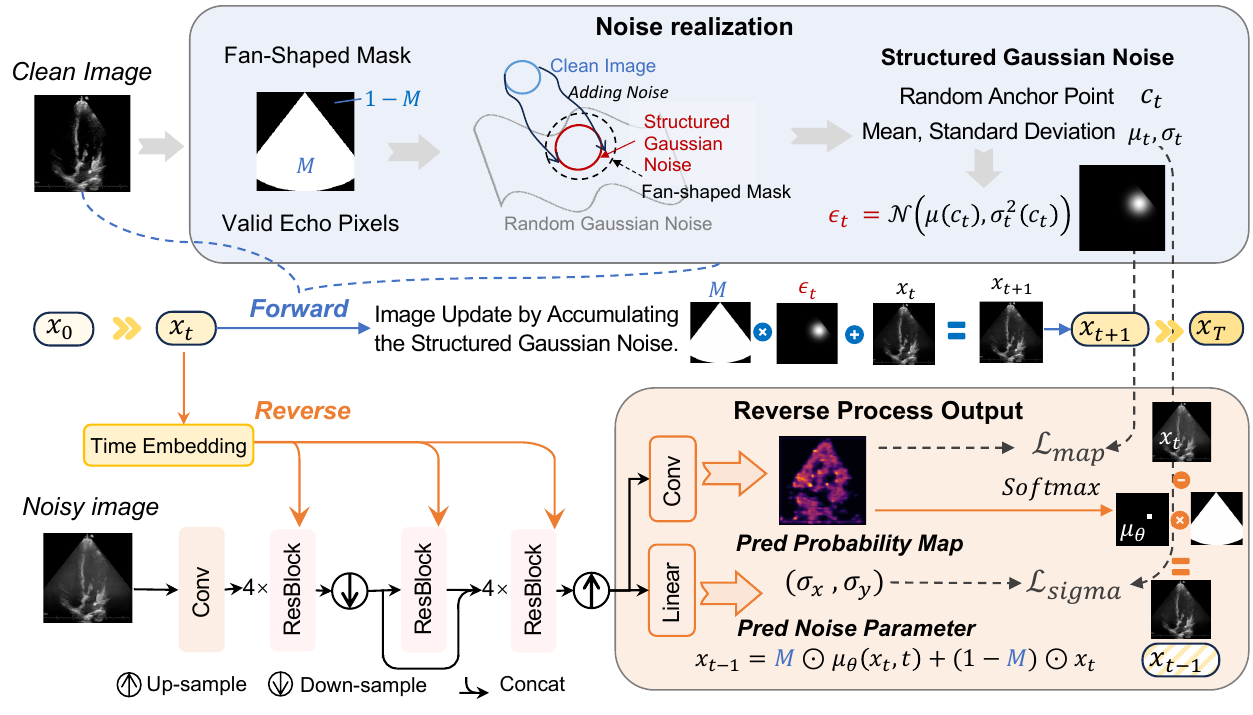}
   \caption{Illustration of the proposed structure-informed diffusion model for echo dehazing. The forward process injects structured Gaussian noise, which is generated as a subset constrained by the fan-shaped mask. This constraint ensures that noise only appears within the valid echo region, unlike random Gaussian noise. The reverse network predicts a probability map and noise parameters using the standard U-Net architecture with two decoder heads. It thus enables iterative removal of structured artifacts and progressive recovery of the clean image.}
   \label{fig:method:framework}
\end{figure*}

\section{Methodology}


\subsection{Forward Process: Structured Gaussian Noise} \label{method:forward}

Echo images are fan-shaped due to transducer scanning geometry, with meaningful anatomical signals confined to the valid fan region. Accordingly, haze artifacts are also restricted to this region.
To incorporate this geometric prior, we first create a binary mask $M$ that identifies valid echo pixels, as shown in Fig. \ref{fig:method:framework}. The global mask is generated by tuning the fan parameters to align accurately with the signal-containing echo region, yielding a mask consistent with ultrasound imaging geometry. 
Given the fan-shaped domain defined by mask $M$, we introduce a stochastic mechanism to generate spatially correlated Gaussian noise that realistically simulates haze artifacts. Rather than adding independent pixel-wise Gaussian noise, our approach creates localized noise patterns whose influence decays radially from randomly selected anchor points within the echo region.
At each diffusion timestep $t \in [1, T]$, we sequentially sample an anchor point, generate structured Gaussian noise, and update the image.

We first sample an anchor point $c_t = (i_t, j_t)$ from the set of valid echo pixels that also satisfy a location constraint parameterized by $\ell \in [0,1]$:
\begin{equation}
c_t \sim \mathcal{C}\left(\{ (i,j) : M(i,j) = 1 \;\text{and}\; j \le \ell \cdot L \}\right),
\end{equation}
where $L$ is the image height and $\ell$ controls the proportion of the image from the top where the anchor point can be sampled. 
We argue a single fixed noise pattern would be too simplistic, as real echo haze varies across images and can appear at different locations within the fan. 
Therefore, we employ stochastic selection on $c_t$ to ensure that haze artifacts can originate from any location within the echo domain, reflecting the variable nature of acoustic multipath interference.
As shown in Fig. \ref{fig:intro:teaser}, by randomizing the center location at each timestep, we effectively create a mixture model that can approximate complex haze distributions:
\begin{equation}
p_H(h) \approx \sum_{t=1}^{T} \mathcal{N}_2\left( \boldsymbol{\mu}_t(c_t), \boldsymbol{\Sigma}_t(\sigma_t) \right) \odot M,
\end{equation}
where $\mathcal{N}_2(\cdot)$ denotes a two-dimensional Gaussian distribution, $\boldsymbol{\mu}_t(c_t)$ is the mean vector (center location) of the Gaussian blob determined by the anchor point $c_t$, and $\odot$ denotes element-wise multiplication. $\boldsymbol{\Sigma}_t(\sigma_t)$ is its covariance matrix controlling the spread and orientation of the haze pattern, which is sampled from a predefined set $\Theta$ via a random parameter $\sigma_t \sim \mathcal{C}(\Theta)$.
This means the overall distribution of haze $p_H(h)$ is represented as a sum of many different Gaussian blobs, each centered at a different random location within the echo fan. This mixture can capture the diverse, spatially varying characteristics of real echo haze much better than a single Gaussian distribution.

Therefore, in the forward stage of the diffusion model, the noise added at timestep $t$ is sampled from a two-dimensional Gaussian distribution parameterized by the spatially varying statistics:
\begin{equation}
h_t(c_t,\sigma_t) \sim \mathcal{N}_2\left( \boldsymbol{\mu}_t(c_t), \boldsymbol{\Sigma}_t(\sigma_t) \right) \odot M.
\end{equation}
The noisy image is updated by accumulating the structured noise:
\begin{equation}
\tilde{x}_t = \sqrt{\alpha_t} \, x_{t-1} + \sqrt{1 - \alpha_t} \, h_t,
\end{equation}
\begin{equation}
{x}_t = \mathcal{H} \odot \tilde{x}_t + (1 - \mathcal{H}) \odot {x}_{t-1},
\label{eq:mask-application}
\end{equation}
where \(\alpha_t\) is the scheduling coefficient, and \(\mathcal{H}\) represents a binary mask that takes the value \(1\) in regions where \(h_t\) is non-zero and \(0\) in regions where \(h_t\) is zero.
The stochastic center sampling across timesteps generates a diverse ensemble of spatially correlated noise realizations. 
This ensemble effectively models the heterogeneous distribution of real haze, which may arise from multiple scattering sources at different locations within the imaging field. The resulting forward process produces hazy images that respect both the geometric constraints of echo imaging and the statistical properties of acoustic clutter.
Unlike standard diffusion where the forward process always ends in pure noise across the entire image (as shown in Fig. \ref{fig:intro:teaser}), our process ends with an image $x_T$ that contains realistic haze only within the echo fan, matching the actual geometry of echo data where corners are already black.

\subsection{Reverse Process: Structured Noise Estimation via Probabilistic Coordinate Prediction} \label{method:reverse}
The reverse process aims to recover the clean image $x_0$ by iteratively removing the structured noise added during the forward process. At each timestep $t$, we must estimate the noise component $h_t$ that was injected. A naive approach would directly predict the anchor point coordinates $c_t$ used in the forward process. However, we observe that direct coordinate regression introduces significant bias, as the model tends to predict coordinates near the image center to minimize the empirical loss, failing to capture the spatial diversity of real haze patterns. We therefore propose a probabilistic mapping strategy that avoids this bias through indirect coordinate estimation.
Let $\epsilon_\theta(x_t, t)$ denote the predicted structured noise at timestep $t$. Rather than directly regressing the anchor point coordinates, our network first estimates a positional probability map $\mathcal{P}_t \in \mathbb{R}^{H \times W}$ that encodes the likelihood of each pixel being the noise center:
\begin{equation}
\mathcal{P}_t = \text{softmax}\left( \Phi_{\text{prob}}(x_t, t) \right),
\end{equation}
where $\Phi_{\text{prob}}$ is an output head of the U-Net \cite{conf/MICCAI/ronneberger2015} that produces a 2D activation map. The predicted anchor point $(\hat{\mu}_x, \hat{\mu}_y)$ is then obtained as the coordinates of the maximum probability:
\begin{equation}
(\hat{\mu}_x, \hat{\mu}_y) = \arg\max_{i,j} \mathcal{P}_t(i,j).
\end{equation}
This probabilistic formulation offers two key advantages. 
First, it naturally models uncertainty in the anchor point location, reflecting the fact that haze artifacts may not have a single well-defined origin. Second, it avoids the centroid bias inherent in direct regression, as the network must learn to produce high-probability peaks at diverse locations across the training distribution.
Note that the argmax operation is used only at inference stage to extract the anchor point coordinates. 

In addition to the probability map, the network also predicts the spatially varying noise parameters  $\hat{\sigma}_{\text{t}}$. Together with the predicted anchor point, these parameters define the estimated noise field via the same radial basis formulation used in the forward process:
\begin{equation}
\hat{h}_t \sim \mathcal{N}_2\left( \boldsymbol{\hat\mu}_t, \boldsymbol{\Sigma}_t(\hat\sigma_t) \right) \odot M.
\end{equation}
For notational convenience, we denote the complete noise prediction as $\epsilon_\theta(x_t, t) = \hat{h}_t$.
Following the denoising diffusion probabilistic model (DDPM)~\cite{conf/NeurIPS/ho2020} formulation, the predicted mean $\mu_\theta(x_t, t)$ for the reverse transition is derived from the estimated noise. Unlike the original stochastic DDPM, we adopt a deterministic sampling strategy by omitting the random noise term, which improves stability and reproducibility:
\begin{equation}
\mu_\theta(x_t, t) = \frac{1}{\sqrt{\alpha_t}} \left( x_t - \frac{1 - \alpha_t}{\sqrt{1 - \bar{\alpha}_t}} \epsilon_\theta(x_t, t) \right),
\label{eq:model_mean}
\end{equation}
where $\bar{\alpha}_t = \prod_{s=1}^{t} \alpha_s$ follows the standard noise schedule. This formulation ensures that $\mu_\theta(x_t, t)$ approximates the less-corrupted image $x_{t-1}$ at the previous timestep.
To respect the geometric constraints of echo imaging, we apply the fan mask during each reverse update. Pixels within the echo domain are updated according to the predicted mean, while pixels outside the domain (the black corners) remain unchanged:
\begin{equation}
\hat{x}_{t-1} = \hat{\mathcal{H}} \odot \mu_\theta(x_t, t) + (1 - \hat{\mathcal{H}}) \odot \hat{x}_t,
\label{eq:update_step}
\end{equation}
where \(\hat{\mathcal{H}}\) represents the predicted binary mask relevant to $\hat{h_t}$. This masking strategy prevents the generation of artifacts in regions without valid echo data.

\subsection{Optimization} \label{method:optimization}

The proposed diffusion model is trained to accurately estimate the structured noise parameters that govern the forward process. We formulate a multi-objective loss function that separately supervises the probabilistic anchor point prediction and the noise distribution parameters, enabling the network to learn both the spatial origins and statistical properties of realistic haze patterns.

\subsubsection{Probabilistic Map Loss}

The positional probability map $\mathcal{P}_t$ encodes the likelihood of each pixel being the anchor point for structured noise at timestep $t$. During training, we have access to the true anchor point $c_t$ sampled during the forward process. We supervise the probability map using a mean squared error loss that minimizes the discrepancy at the ground-truth anchor point location:
\begin{equation}
\mathcal{L}_{\text{map}} = \mathbb{E}_{t, x_t, c_t} \left[ \|h_t - \mathcal{P}_t\|^2 \right].
\label{eq:map_loss}
\end{equation}
This formulation encourages the network to assign high probability to the actual noise origin while naturally modeling uncertainty through the softmax distribution. Unlike direct coordinate regression, which would penalize all off-center predictions equally, our probabilistic approach allows the network to maintain multi-modal hypotheses about potential noise origins, better reflecting the ambiguous nature of haze sources.

\subsubsection{Noise Parameter Regression Loss}

Accurate estimation of the spatially varying noise parameters is essential for faithful haze simulation and removal. We supervise the predicted noise field parameters against their ground-truth values used in the forward process:
\begin{equation}
\mathcal{L}_{\text{sigma}} = \| \hat{\sigma}_t - \sigma_t \|^2.
\end{equation}
This direct parameter regression ensures that the network learns the correct scale and spatial extent of the noise patterns. 

\subsubsection{Total Loss}

The complete training objective combines the two loss components with appropriate weighting:

\begin{equation}
\mathcal{L}_{\text{total}} = \mathcal{L}_{\text{map}} + \lambda \mathcal{L}_{\text{sigma}},
\label{eq:method:loss}
\end{equation}
where $\lambda$ is a balancing parameter.

\section{Experiments and Results} 

\subsection{Dataset and Pre-Processing} \label{exp:data}

We evaluated our method on the public dataset from DehazingEcho2025 Challenge \footnote{\url{https://dehazingecho2025.grand-challenge.org/}}, which contains 4,500 four-chamber echo images from 75 easy-to-image patients and 2,400 from 40 difficult-to-image patients. 
For clean images, the easy-to-image cases were split with a $9:1$ ratio, allocating $90\%$ for training and the remainder for validation and test. For noisy images, the challenge provided 237 masked images used for validation and test. 
The masks defined two specific anatomical regions: the apical heart chamber (left ventricle), which typically presents the most challenging and haze-affected area, as Region A, and a segment of the ventricular septum (wall) as Region B. 
Note that here Regions A and B are defined based on the haze-affected areas within the left ventricle and septal wall, and do not correspond exactly to the anatomical structures themselves.
All images were converted to grayscale and augmented with random flipping during preprocessing.

For external validation, we used two internal echo datasets acquired at the National University Hospital (NUH), Singapore: NUH-HCM and NUH-MI, comprising images from patients of hypertrophic cardiomyopathy (HCM) and post myocardial infarction (MI), respectively. To identify noisy apical four-chamber (A4C) frames, we trained a classifier on the data provided by DehazingEcho2025 and applied the frozen model to A4C images from both NUH datasets. Frames classified as noisy were selected for dehazing evaluation, yielding 122 frames from NUH-HCM and 735 frames from NUH-MI. For NUH-MI, evaluation masks were generated by a segmentation model pretrained on the DehazingEcho2025 dataset using the provided region masks. For NUH-HCM, we used manually annotated labels as evaluation masks, defining the left ventricle as Region A and the myocardium as Region B instead.

\subsection{Evaluation Metric} \label{exp:label}

We used generalized Contrast-to-Noise Ratio (gCNR), the tissue edge preservation index ($\beta$/EPI), and Fr\'echet Inception Distance (FID) for model validation.
gCNR measures the degree of histogram overlap between a target region and the background:
\begin{equation}
\text{gCNR} = 1 - \int \min\{p_A(x), p_B(x)\} \, dx,
\label{eq:gcnr}
\end{equation}
where $p_A(x)$ and $p_B(x)$ represents the distribution of Regions A and B, respectively. 
A larger gCNR value indicates better contrast separation between the anatomical structure of interest and the surrounding tissue after dehazing.
$\beta$/EPI assesses edge preservation within the tissue region (Region B), with higher values indicating better structural preservation.
FID compares the feature distributions of dehazed and clean reference image sets:
\begin{equation}
\mathrm{FID}=\|\mu_d-\mu_c\|_2^2+\operatorname{Tr}\!\left(\Sigma_d+\Sigma_c-2(\Sigma_d\Sigma_c)^{1/2}\right),
\label{eq:fid}
\end{equation}
where $\mu$ and $\Sigma$ denote the mean and covariance of Inception features; subscripts $d$ and $c$ refer to dehazed and clean reference image sets. A lower FID indicates closer agreement. 

\subsection{Implementation Details}

All experiments were conducted on an NVIDIA RTX 4090 GPU with 24 GB memory. 
We employed a cosine noise schedule for the forward diffusion process and optimized the network using the Adam optimizer with gradient clipping to ensure stable convergence.
The batch size was set to 4, and the model was trained for 100 epochs with an initial learning rate of $5\times10^{-5}$.
In Eq. \ref{eq:method:loss}, instead of fixing $\lambda$, we employed an adaptive loss balancing strategy. Specifically, the gradient flow from the larger loss term was dynamically scaled to match that of the smaller term by detaching the ratio between them. This stabilization technique prevented the optimization from being dominated by a single component without introducing additional learnable parameters. Specifically, the U-Net architecture was configured with a base channel number of 64, channel multipliers of $[1,2,4,8]$ across four resolution stages, and each stage comprised 4 residual blocks.

\begin{figure}[t]
  \centering
   \includegraphics[width=1\linewidth]{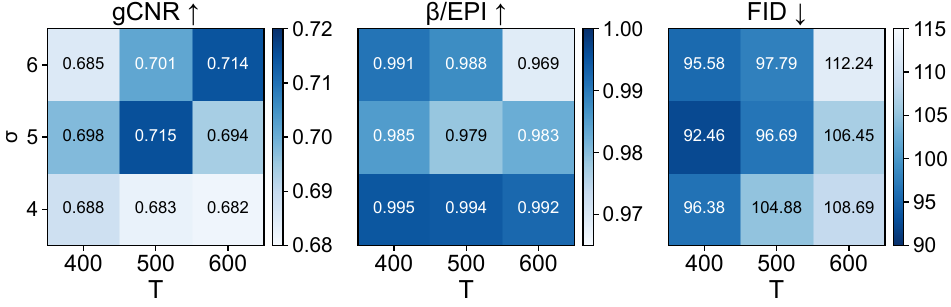}
   \caption{Performance of the model under different parameter combinations. $T$ refers the total number of diffusion steps, while $\sigma$ is the standard deviation of the added Gaussian noise in the forward process.}   
   \label{fig:ps:heatmap}
\end{figure}

\subsection{Parameter Study}

We conducted a parameter study to examine the influence of the total number of diffusion steps $T$, the standard deviation $\sigma$ of the Gaussian component, and the spatial extent $\ell$ of noise injection. 
As shown in Fig.~\ref{fig:ps:heatmap}, at $T=500$, increasing $\sigma$ from 4 to 5 improved gCNR from 0.683 to 0.715 and reduced FID from 104.88 to 96.69. This setting achieved the highest gCNR, with a tissue $\beta$/EPI of 0.979. At $T=600$ and $\sigma=6$, gCNR remained similar (0.714), while $\beta$/EPI decreased to 0.969 and FID increased to 112.24. These results support $T=500$ and $\sigma=5$ as a balanced setting for haze removal and structural preservation.

We further investigated the spatial extent of noise injection, as summarized in Table~\ref{tab:exp:PS_location_results}. Restricting injection to $\ell=0.3$ or $0.5$ yielded gCNR values of 0.659--0.660, whereas full-region injection ($\ell=1.0$) achieved the highest gCNR of $0.715 \pm 0.090$ and the lowest FID of 96.69, with a tissue $\beta$/EPI of $0.979 \pm 0.016$. We therefore adopted full-region injection to capture the spatial variability of echo haze.

\begin{table}[t]
\centering
\caption{Model performance under different noise injection ranges. Here, $\ell$ represents the proportion of the image height (from the top) where structured noise is applied.}
\label{tab:exp:PS_location_results}
\resizebox{0.7\linewidth}{!}{
\begin{tabular}{l c c c}
    \toprule
    $\ell$ & gCNR $\uparrow$ & $\beta$/EPI $\uparrow$ & FID $\downarrow$ \\
    \midrule
    0.3 & $0.660 \pm 0.090$ & $0.979 \pm 0.019$ & $105.54$ \\
    0.5 & $0.659 \pm 0.084$ & $0.988 \pm 0.009$ & $107.59$ \\
    0.7 & $0.650 \pm 0.167$ & $0.972 \pm 0.034$ & $105.52$ \\
    0.9 & $0.693 \pm 0.092$ & \bm{$0.992 \pm 0.007$} & $98.49$ \\
    1.0 & \bm{$0.715 \pm 0.090$} & $0.979 \pm 0.016$ & \bm{$96.69$} \\
    \bottomrule
\end{tabular}}
\end{table}

\subsection{Comparison Study}  

\begin{table}[t]
  \caption{Quantitative comparison of dehazing methods on the DehazingEcho2025 dataset. } 
  \label{tab:exp:results}
  \centering
   \resizebox{0.9\linewidth}{!}{
  \begin{tabular}{{llll}}
    \toprule
    Method & gCNR $\uparrow$ & $\beta$/EPI $\uparrow$ & FID $\downarrow$ \\
    \midrule
    BM3D~\cite{conf/SPIE/Dabov2006} & $0.637 \pm 0.130$ & $0.396 \pm 0.111$ & $285.86$ \\
    CycleGAN~\cite{conf/ICCV/zhu2017} & $0.651 \pm 0.111$ & $0.794 \pm 0.038$ & $97.49$ \\
    DDPM~\cite{conf/NeurIPS/ho2020} & $0.650 \pm 0.147$ & $0.188 \pm 0.116$ & $268.15$ \\
    RCOT~\cite{conf/CVPR/tang2024} & $0.679 \pm 0.110$ & \bm{$0.997 \pm 0.003$} & $100.74$ \\
    CycleDiff~\cite{journal/TIP/zou2026} & $0.675 \pm 0.150$ & $0.257 \pm 0.077$ & $147.72$ \\
    FrDiff~\cite{conf/CVPR/liu2025} & $0.689 \pm 0.144$ & $0.871 \pm 0.033$ & $101.91$ \\
    EchoNet-Quality~\cite{Choi_2025_ICCV} & $0.681 \pm 0.128$ & $0.948 \pm 0.035$ & \bm{$94.68$} \\
    \textbf{Ours} & \bm{$0.715 \pm 0.090$} & $0.979 \pm 0.016$ & $96.69$ \\
    \bottomrule
  \end{tabular}}
\end{table}

\begin{figure*}[t]
  \centering
   \includegraphics[width=0.98\linewidth]{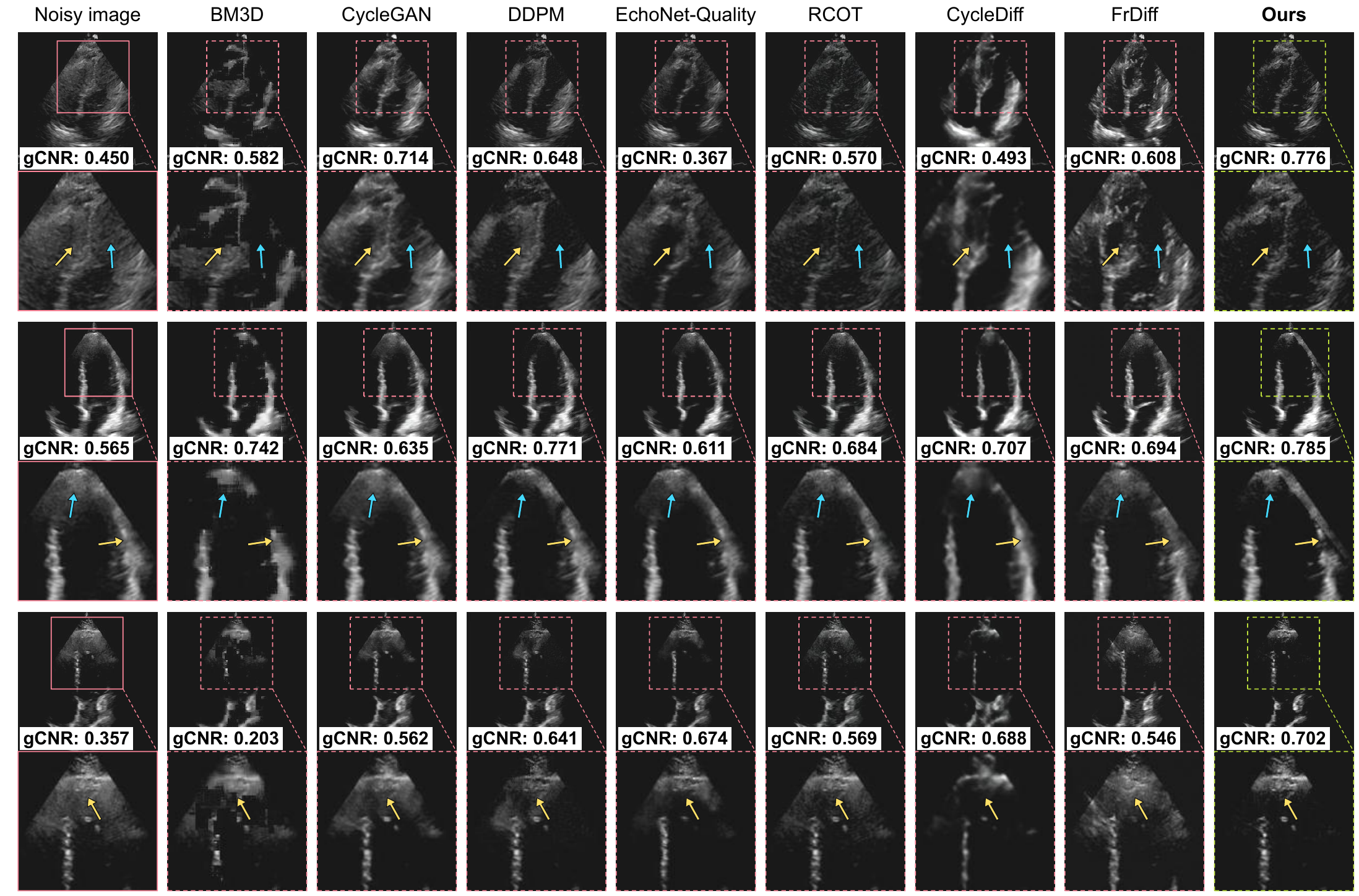}
   \caption{Comparison of echo dehazing performance across different algorithms. The  arrows in the enlarged box highlights the area for visual comparison, and gCNR is reported for each image.}
   \label{fig:exp:comparison}
\end{figure*}

Table~\ref{tab:exp:results} presents the quantitative comparison on the DehazingEcho2025 dataset. We compared our method with BM3D~\cite{conf/SPIE/Dabov2006} for conventional denoising, CycleGAN~\cite{conf/ICCV/zhu2017} and CycleDiff~\cite{journal/TIP/zou2026} for unpaired image-to-image translation, standard DDPM~\cite{conf/NeurIPS/ho2020}, RCOT~\cite{conf/CVPR/tang2024} for structure-preserving restoration, FrDiff~\cite{conf/CVPR/liu2025} for frequency-domain dehazing, and EchoNet-Quality~\cite{Choi_2025_ICCV}, an ultrasound-specific generative model. Our method achieved the highest mean gCNR of $0.715 \pm 0.090$, indicating improved contrast separation between cardiac chambers and surrounding tissue. This result supported the benefit of explicitly representing spatially correlated haze through localized Gaussian components, allowing the model to target structured artifacts rather than relying solely on generic denoising or image distribution alignment. Our method also achieved a tissue $\beta$/EPI of $0.979 \pm 0.016$, second only to RCOT ($0.997 \pm 0.003$). However, RCOT achieved a lower gCNR of $0.679 \pm 0.110$, suggesting that strong edge preservation did not necessarily translate into effective haze suppression. Similarly, EchoNet-Quality achieved the lowest FID (94.68), but lower gCNR and $\beta$/EPI than our method, illustrating that agreement with the clean image distribution alone did not ensure comparable contrast enhancement and structural preservation. With the second-lowest FID of 96.69, our method provided a favorable balance across these complementary criteria.

Figure~\ref{fig:exp:comparison} presents visual comparisons on three randomly selected subjects. Conventional denoising with BM3D altered image texture and introduced block-like artifacts without consistently removing the spatially extended haze. Generative translation methods reduced haze in some regions but also smoothed myocardial texture and altered local boundary appearance, as illustrated by CycleGAN and CycleDiff at the arrow-marked locations. DDPM, EchoNet-Quality, and FrDiff retained varying degrees of diffuse haze or granular artifacts, while RCOT preserved tissue detail but left residual intracavitary haze. These observations suggest that generic noise modeling or image translation objectives may not adequately distinguish structured haze from adjacent anatomical signals. By constraining artifact estimation to localized, spatially coherent components, our method reduced intracavitary haze while retaining surrounding tissue boundaries. The yellow and cyan arrows in the first two cases illustrate this balance between clearer chamber interiors and preserved wall definition, while the yellow arrows in the third case highlight reduced near-field haze adjacent to bright tissue. 


\subsection{Ablation Study} \label{sec:exp:ablation}

\begin{table}[t]
\centering
\caption{Comparison of pixel-wise dense versus Gaussian-parameterized noise prediction for structured noise modeling.}
\label{tab:exp:ablation_pred}
\resizebox{0.94\linewidth}{!}{
\begin{tabular}{l c c c}
\toprule
Method & gCNR $\uparrow$ & $\beta$/EPI $\uparrow$ & FID $\downarrow$ \\
\midrule
Ours w/o Gaussian Pre & $0.244 \pm 0.047$ & $0.424 \pm 0.058$ & $280.96$ \\
Ours w/o Fan Mask & $0.681 \pm 0.106$ & \bm{$0.991 \pm 0.006$} & $100.41$ \\
\textbf{Ours} & \bm{$0.715 \pm 0.090$} & $0.979 \pm 0.016$ & \bm{$96.69$} \\
\bottomrule
\end{tabular}}
\end{table}

We examined how the noise representation affected contrast recovery and tissue preservation, as shown in Table~\ref{tab:exp:ablation_pred}.
One can see that replacing direct pixel-wise noise prediction with Gaussian parameterization improved all three metrics. 
The concurrent gains in gCNR and tissue $\beta$/EPI suggest that the structured representation improved chamber-tissue separability while preserving tissue edges. By estimating the parameters of localized Gaussian components, the reverse process modeled haze as a spatially coherent intensity pattern. We also varied the number of components introduced and predicted per diffusion step while keeping the other settings fixed (Table~\ref{tab:exp:ablation_number}). A single component achieved the best performance across all three metrics. Although additional components increased the representational capacity of each step, localized components accumulated across successive steps, allowing the forward process to represent spatially heterogeneous haze with one component per step.
We also assessed the role of the fan mask by removing it from both the forward and reverse processes, which resulted in reduced gCNR and increased FID, while slightly improving tissue $\beta$/EPI (Table~\ref{tab:exp:ablation_pred}).
This trade-off suggests that the main benefit of the fan constraint was improved chamber-tissue separability and closer agreement with the reference image distribution, with tissue edge preservation remaining high in both settings. A plausible explanation is that restricting anchors to the ultrasound sector concentrated the modeled degradation within the region containing valid echo signal.

\begin{table}[t]
\centering
\caption{Impact of the number of Gaussian components.}
\label{tab:exp:ablation_number}
\resizebox{0.7\linewidth}{!}{
\begin{tabular}{l c c c}
\toprule
Num & gCNR $\uparrow$ & $\beta$/EPI $\uparrow$ & FID $\downarrow$ \\
\midrule
1 & \bm{$0.715 \pm 0.090$} & \bm{$0.979 \pm 0.016$} & \bm{$96.69$} \\
2 & $0.708 \pm 0.116$ & $0.970 \pm 0.020$ & $108.08$ \\
3 & $0.679 \pm 0.129$ & $0.974 \pm 0.043$ & $119.06$ \\
4 & $0.681 \pm 0.156$ & $0.968 \pm 0.038$ & $133.96$ \\
5 & $0.692 \pm 0.164$ & $0.931 \pm 0.082$ & $156.55$ \\
\bottomrule
\end{tabular}}
\end{table}

\begin{figure}[t] 
  \centering
   \includegraphics[width=0.7\linewidth]{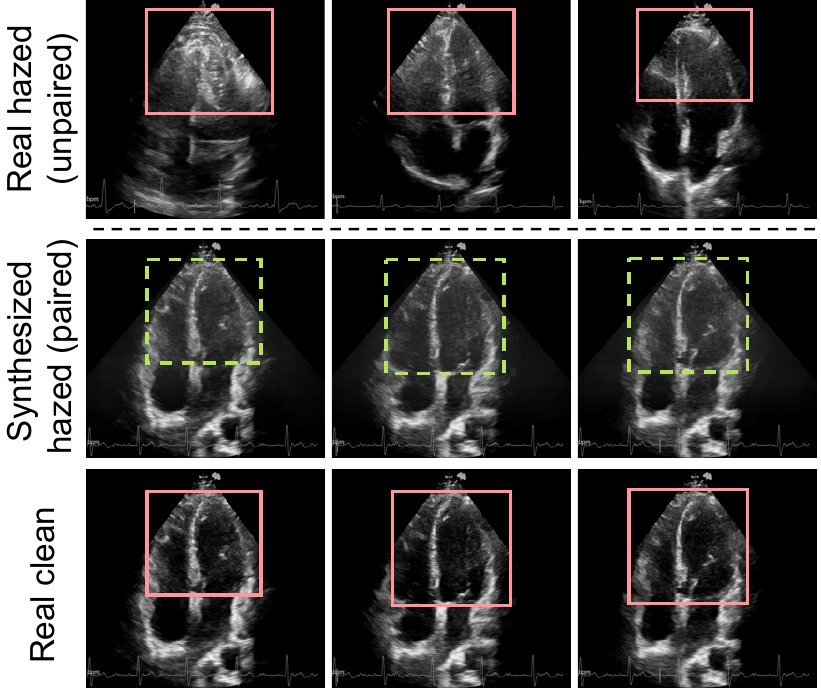}
   \caption{ Illustration of generated realistic haze distribution in the forward stage of the proposed method.}
   \label{fig:result:anchor}
\end{figure}

\begin{figure}[t]
  \centering
  \includegraphics[width=0.8\linewidth]{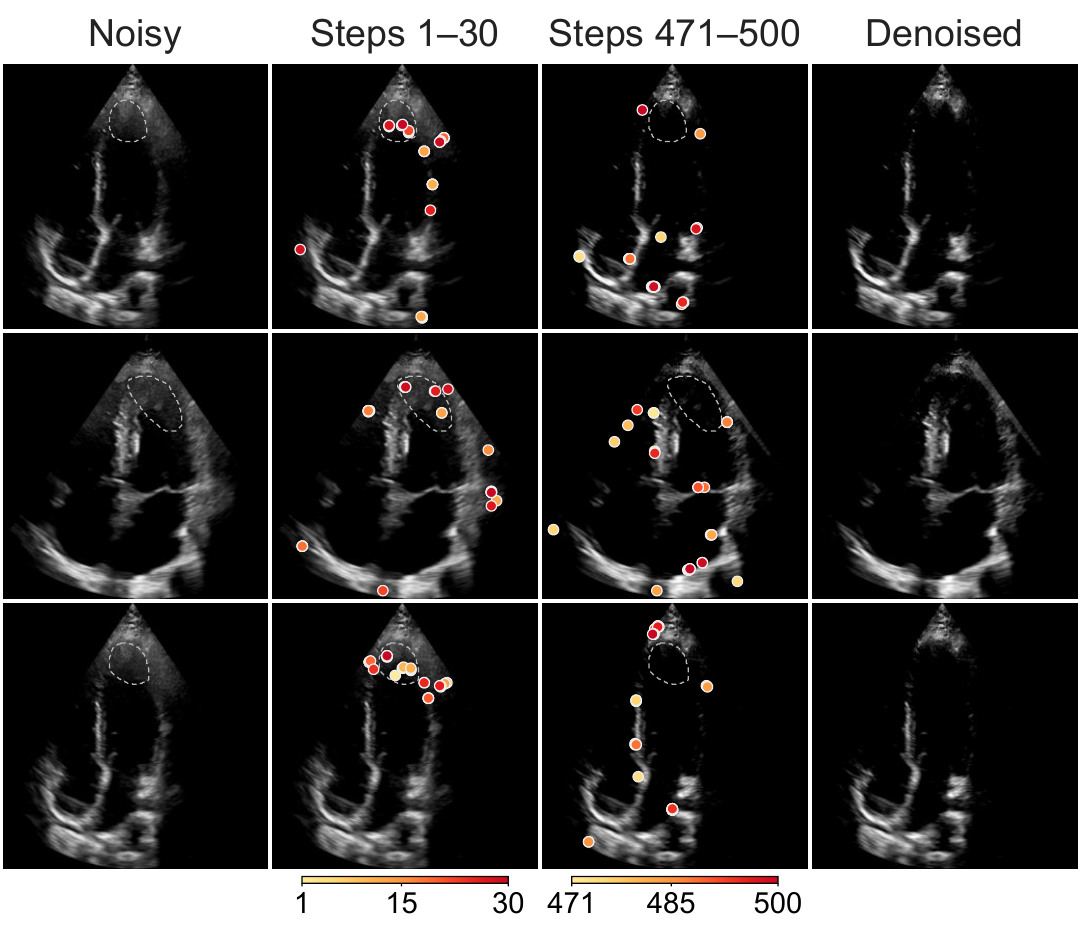}
  \caption{Evolution of predicted anchor locations during the reverse diffusion. From left to right, each row shows the hazy input, intermediate images overlaid with anchors from early (steps 1--30) and late (steps 471--500) inference, and the final dehazed image. Dashed contours delineate regions of pronounced near-field haze. Each dot marks a predicted anchor, with its color indicating the inference step.}
  \label{fig:anchor_distribution}
\end{figure}

\subsection{Evaluation of Realistic Haze Modeling and Anchor Diversity}

We examined whether the proposed forward process reproduced the spatial appearance of echo haze and whether the reverse process adapted its predicted anchors to different artifact locations. Fig.~\ref{fig:result:anchor} presents unpaired real hazy images as qualitative references in the first row, followed by synthesized hazy images and their corresponding clean inputs. Within the highlighted near-field regions, the synthesized haze produced spatially varying intensity elevation, reduced chamber-tissue contrast, and obscured anatomical details that were visible in the clean inputs. These changes resembled the localized loss of visibility observed in the real hazy examples. Rather than uniformly perturbing the image, the accumulated Gaussian components generated spatially coherent degradation with varying local extent. This qualitative agreement supported the suitability of localized Gaussian components as an approximation of the spatial appearance of echo haze, although it did not establish equivalence to the underlying acoustic formation process.

We further investigated how the predicted anchor locations evolved during reverse diffusion, as shown in Fig.~\ref{fig:anchor_distribution}. During the first 30 inference steps, anchors frequently clustered within or near regions of pronounced near-field haze, suggesting that the model initially concentrated artifact removal on areas with substantial degradation. During the final inference steps (471-500), the ventricular cavities appeared clearer, while anchors were more dispersed across the ultrasound sector, including near tissue boundaries. This redistribution was consistent with a transition from suppressing prominent localized haze to making more spatially distributed corrections. The variation in anchor locations across images and inference stages also suggested that the probability-map formulation accommodated image-dependent artifact patterns rather than repeatedly selecting a fixed central location.

\begin{figure}[t]
\centering
\includegraphics[width=0.8\linewidth]{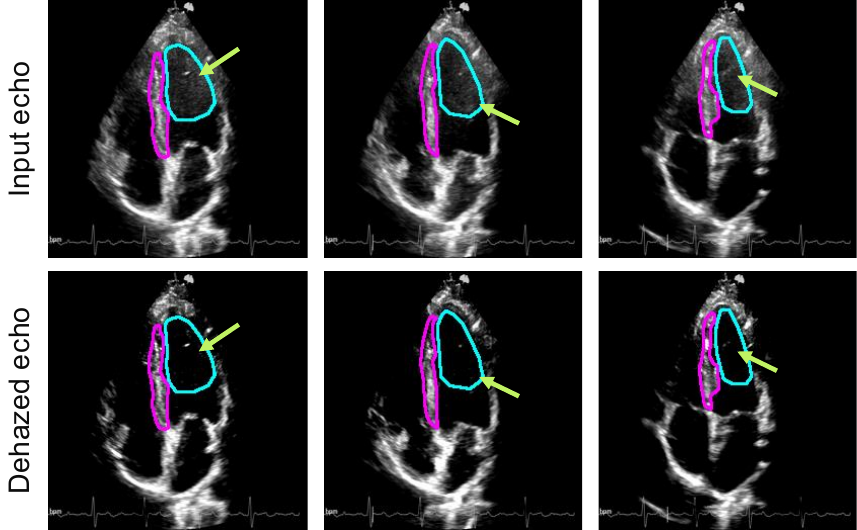}
\caption{Illustration of input and denoised image overlapped with contours of Regions A (cyan) and B (magenta). The arrows indicate the masked region where haze has been effectively removed.}
\label{fig:exp:dehazing}
\end{figure}

\begin{table}[t]
  \caption{Comparison of downstream segmentation task performance before and after echo dehazing.}
  \label{tab:downsteam result}
  \centering
  \resizebox{0.94\linewidth}{!}{
  \begin{tabular}{{l c c c}}
    \toprule
    Method & Dice $\uparrow$ & IoU $\uparrow$ \\
    \midrule
    Without Processing (Original Image) & {$0.805\pm 0.134$} & {$0.688\pm 0.135$} \\
    EchoNet-Quality~\cite{Choi_2025_ICCV} & {$0.824\pm 0.103$} & {$0.713\pm 0.112$}  \\
    \textbf{Ours} & \bm{$0.844\pm 0.091$} & \bm{$0.746\pm 0.102$} \\
    \bottomrule
  \end{tabular}}
\end{table}

\subsection{Downstream Anatomical Delineation of Dehazed Echo}

Figure~\ref{fig:exp:dehazing} shows representative denoising results with mask overlays. The method effectively suppressed localized haze, such as in the apical chamber, while preserving fine structures like the septal wall.
To assess the practical utility of our dehazing method, we evaluated U-Net~\cite{conf/MICCAI/ronneberger2015} segmentation of the ventricular septum on original and dehazed images. The ventricular septum was chosen for its well-defined anatomical position and susceptibility to haze artifacts from acoustic interference. As summarized in Table \ref{tab:downsteam result}, our method outperformed both the original images and the state-of-the-art EchoNet-Quality model, achieving a Dice score of $0.844 \pm 0.091$ and IoU of $0.746 \pm 0.102$, highlighting superior boundary precision and regional consistency. These results suggest that our dehazing approach improved image quality while providing modest gains in downstream segmentation accuracy.

\begin{table}[t]
    \caption{External validation on two internal datasets.}
    \label{tab:external_validation}
    \centering
    \resizebox{0.99\linewidth}{!}{
    \begin{tabular}{llccc}
    \toprule
    Dataset & Method & gCNR $\uparrow$ & $\beta$/EPI $\uparrow$ & FID $\downarrow$ \\
    \midrule
    \multirow{7}{*}{NUH-HCM}
     & BM3D~\cite{conf/SPIE/Dabov2006} & $0.596 \pm 0.180$ & $0.361 \pm 0.100$ & $207.74$ \\
     & CycleGAN~\cite{conf/ICCV/zhu2017} & $0.673 \pm 0.219$ & $0.724 \pm 0.043$ & $169.95$ \\
     & DDPM~\cite{conf/NeurIPS/ho2020} & $0.465 \pm 0.144$ & $0.100 \pm 0.073$ & $348.95$ \\
     & RCOT~\cite{conf/CVPR/tang2024} & $0.660 \pm 0.156$ & \bm{$0.997 \pm 0.003$} & $118.94$ \\
     & CycleDiff~\cite{journal/TIP/zou2026} & $0.597 \pm 0.153$ & $0.135 \pm 0.034$ & $212.89$ \\
     & FrDiff~\cite{conf/CVPR/liu2025} & $0.639 \pm 0.176$ & $0.913 \pm 0.024$ & $169.90$ \\
     & EchoNet-Quality~\cite{Choi_2025_ICCV} & $0.691 \pm 0.195$ & $0.943 \pm 0.019$ & $159.50$ \\
     & \textbf{Ours} & \bm{$0.701 \pm 0.148$} & $0.993 \pm 0.010$ & \bm{$112.74$} \\
    \midrule
    \multirow{7}{*}{NUH-MI}
     & BM3D~\cite{conf/SPIE/Dabov2006} & $0.757 \pm 0.128$ & $0.443 \pm 0.113$ & $153.85$ \\
     & CycleGAN~\cite{conf/ICCV/zhu2017} & $0.781 \pm 0.113$ & $0.729 \pm 0.043$ & $82.46$ \\
     & DDPM~\cite{conf/NeurIPS/ho2020} & $0.687 \pm 0.141$ & $0.123 \pm 0.106$ & $367.89$ \\
     & RCOT~\cite{conf/CVPR/tang2024} & $0.782 \pm 0.100$ & \bm{$0.999 \pm 0.001$} & \bm{$49.70$} \\
     & CycleDiff~\cite{journal/TIP/zou2026} & $0.782 \pm 0.127$ & $0.168 \pm 0.078$ & $184.22$ \\
     & FrDiff~\cite{conf/CVPR/liu2025} & $0.777 \pm 0.105$ & $0.927 \pm 0.017$ & $67.90$ \\
     & EchoNet-Quality~\cite{Choi_2025_ICCV} & $0.775 \pm 0.108$ & $0.950 \pm 0.023$ & $64.68$ \\
     & \textbf{Ours} & \bm{$0.785 \pm 0.100$} & \bm{$0.999 \pm 0.002$} & $50.18$ \\
    \bottomrule
    \end{tabular}}
\end{table}

\begin{figure*}[t]
  \centering
  \includegraphics[width=\linewidth]{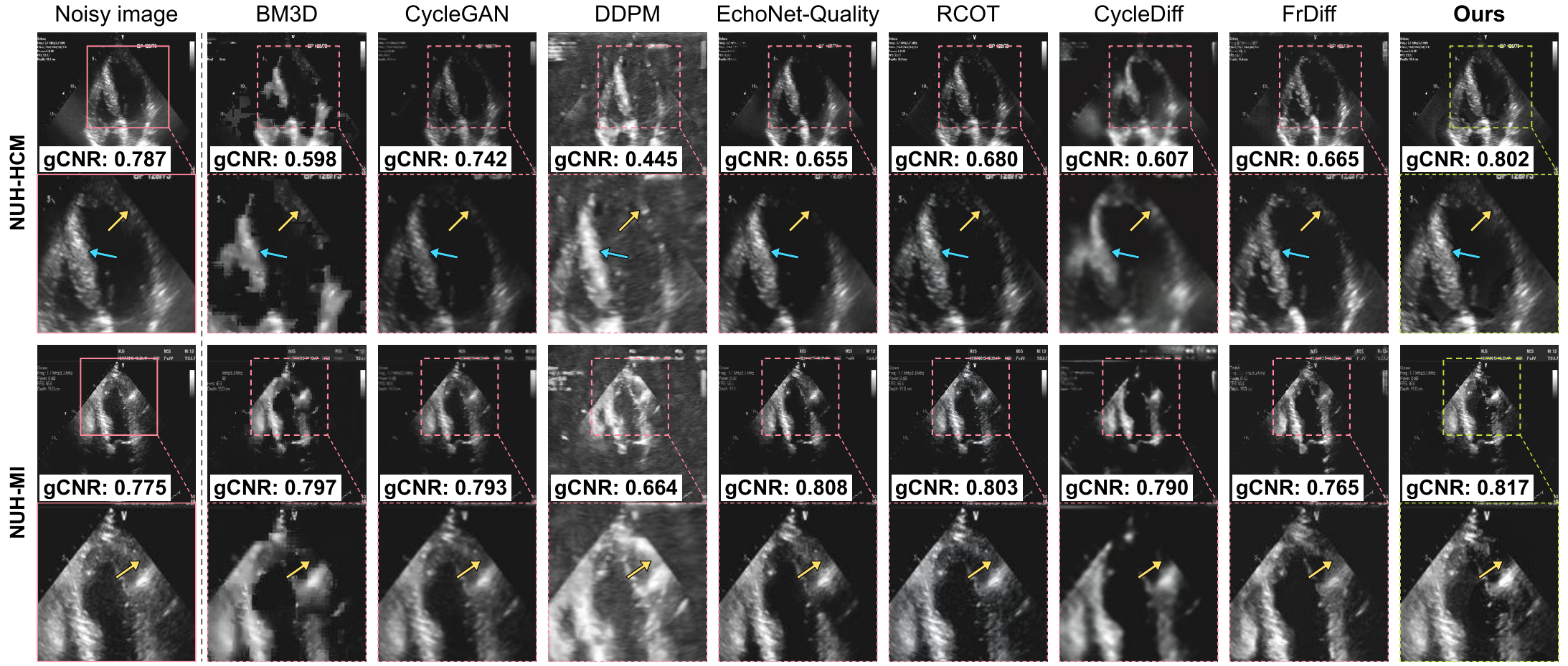}
  \caption{Visual comparison of dehazing results on two internal datasets, i.e., NUH-HCM (top) and NUH-MI (bottom). }
  \label{fig:external:comparison}
\end{figure*}



\subsection{External Validation} \label{subsec:external_validation}

To assess generalization beyond DehazingEcho2025, we compared our method with seven baselines on NUH-HCM and NUH-MI without retraining or fine-tuning on either cohort. 
As shown in Table~\ref{tab:external_validation}, our method achieved the best overall balance across the three metrics, with the highest mean gCNR on both cohorts, the lowest FID on NUH-HCM, and tissue $\beta$/EPI close to or tied with the best baseline.
RCOT~\cite{conf/CVPR/tang2024} showed competitive performance in terms of tissue $\beta$/EPI and FID. Its residual-conditioned refinement may help preserve tissue edges, while its optimal transport objective aligns restored and clean image distributions, consistent with its low FID. However, RCOT achieved lower gCNR than our method on both cohorts, with a larger gap on NUH-HCM (0.660 vs.\ 0.701) than on NUH-MI (0.782 vs.\ 0.785).
Fig.~\ref{fig:external:comparison} further illustrates how the gCNR gains manifested in representative cases. In the boxed regions, our method reduced visible intracavitary haze while retaining the surrounding tissue boundaries. Several baselines left residual haze or altered the appearance of adjacent tissue. 

The NUH-HCM and NUH-MI datasets included echo videos, allowing us to assess interframe consistency as a complement to the primary evaluation of single-frame dehazing performance. The trained model dehazed each frame independently, and temporal warping error (TWE) was used to characterize the resulting sequences. We computed TWE before and after dehazing using the same optical flow estimated from the input sequence and the same valid regions. Lower TWE indicated closer agreement between adjacent frames after warping. We compared patient-level TWE using two-sided paired tests with Benjamini--Hochberg correction across temporal tests within each cohort.
TWE increased after dehazing, with a statistically significant increase on NUH-MI ($p_{adj}<0.01$) and a nonsignificant increase on NUH-HCM ($p_{adj}=0.125$). It may be attributed to the regional wall-motion abnormalities after MI, which may make warping error more sensitive to local changes in intensity and boundary visibility after dehazing.


\section{Discussion and Conclusion}

This study introduced a structure-informed diffusion framework that explicitly represented echo haze as spatially correlated, localized artifacts. By combining fan-constrained Gaussian corruption with probabilistic anchor estimation, the framework incorporated ultrasound-specific spatial priors into both artifact simulation and removal. The experimental results supported the benefit of this formulation: our method achieved the highest mean gCNR on DehazingEcho2025 while maintaining competitive tissue edge preservation and FID. Ablation experiments further showed that Gaussian-parameterized prediction outperformed dense noise prediction, while the fan constraint improved contrast separation and agreement with the clean reference distribution. These findings suggested that explicitly constraining the spatial structure of estimated artifacts facilitated haze removal without requiring unconstrained pixel-wise reconstruction. The qualitative resemblance between synthesized and real haze, together with image-dependent anchor distributions, provided additional support for the proposed representation. Furthermore, validation on NUH-HCM and NUH-MI without fine-tuning demonstrated its applicability to independent clinical cohorts, and the improvement in ventricular septum segmentation indicated benefits for downstream anatomical delineation.

Several limitations should be acknowledged. First, localized Gaussian components provided a simplified approximation of haze appearance rather than an explicit model of acoustic reverberation. Complex effects arising from wave propagation, depth-dependent attenuation, and tissue-dependent scattering were not directly represented. Second, the framework processed individual frames independently and did not enforce temporal consistency. The significant increase in TWE on NUH-MI indicated that improvements in single-frame contrast and structural visibility did not necessarily translate into improved interframe agreement. Third, although the private cohorts included distinct cardiac conditions, both were acquired at the same institution, and differences in evaluation-mask generation limited direct comparison across datasets. Finally, downstream evaluation was restricted to ventricular septum segmentation; the effects of dehazing on cardiac measurements and clinical interpretation remained to be established.

Future work will extend the framework to echo sequences by incorporating temporal information into structured artifact estimation and removal. Integrating acquisition-related priors, including depth-dependent attenuation and probe characteristics, may further improve the fidelity of haze modeling. Future studies will also investigate uncertainty-aware and adaptive multi-scale noise representations for challenging low-quality scans. Validation across institutions, imaging systems, and echocardiographic views will be pursued using standardized evaluation protocols. Further studies will also assess the effects of dehazing on cardiac quantification and expert interpretation to determine its utility in clinical workflows.
    
\bibliographystyle{ieeetr}
\bibliography{A_tmi_refs}

\end{document}